\documentclass{article}
\usepackage{ijcai21}

\usepackage{times}

\usepackage{soul}
\usepackage{url}
\usepackage[hidelinks]{hyperref}
\usepackage[utf8]{inputenc}
\usepackage[small]{caption}
\usepackage{graphicx}
\usepackage{amsmath}
\usepackage{booktabs}
\title{Genetic Algorithms For Extractive Summarization}

\author{
William Chen$^1$\and
Kensal Ramos$^1$\and
Kalyan Naidu Mullaguri$^1$ \and
Annie S. Wu$^1$
\\
\affiliations
$^1$University of Central Florida\\

\emails
\{wchen6255, kensalramos, kalyan.mullaguri\}@knights.ucf.edu
\\aswu@cs.ucf.edu
}

\begin{document}

\maketitle

\begin{abstract}
Most current work in NLP utilizes deep learning, which requires a lot of training data and computational power. 
This paper investigates the strengths of Genetic Algorithms (GAs) for extractive summarization, as we hypothesized that GAs could construct more efficient solutions for the summarization task due to their relative customizability relative to deep learning models. This is done by building a vocabulary set, the words of which are represented as an array of weights, and optimizing those set of weights with the GA. These weights can be used to build an overall weighting of a sentence, which can then be passed to some threshold for extraction. Our results showed that the GA was able to learn a weight representation that could filter out excessive vocabulary and thus dictate sentence importance based on common English words.
 
\end{abstract}

\section{Introduction}
In Natural Language Processing (NLP), text summarization can split broadly into two categories: abstractive and extractive. Abstractive summariziation involves producing a summary of central ideas through language understanding. Extractive summarization deals with identifying and isolating the most relevant or important parts of the text. This paper focuses on the latter category, as we hypothesize the strengths of genetic algorithms fit well with the extractive summarization task.

Genetic algorithms \cite{Holland1984} are an evolutionary technique typically used in optimization problems. The Genetic Algorithm (GA) initializes a population of random solution representations for some problem of interest. These candidate solutions are evaluated with a fitness function, which affects if they are able to make it to the next generation of the population. Individuals are "mated" through a process called crossover, which is intended to hopefully combine the best traits of two parents into their children. Random mutation of individuals in the population may also occur to encourage further diversity. The idea is that the solution population will co-evolve and become more fit over time, producing solutions to the problem that are more optimal than the initial population.

\section{Related Work}
Most current work in NLP utilizes deep learning. Recurrent Neural Networks and Long Short-term Memory \cite{lstm} are two of the architectures traditionally favored for sequence-to-sequence problems such as text summarization. Recent breakthroughs with the Transformer \cite{transformer} model architecture have pivoted the field towards that direction, leaving alternatives (especially those outside of the deep learning realm) relatively unexplored. 

\cite{Silla:auto} utilized genetic algorithms for the text summarization task. Their approach did not involve using GAs directly for text summarization, but rather for attribute selection to aid machine learning algorithms in the task. \cite{Liu:summarization} was the first (to our knowledge) to directly apply GAs to text summarization. They utilized both word-level and sentence-level statistics to guide summarization, achieving near state-of-the-art results on the DUC2004 dataset.

Further work was done by \cite{fattah2008automatic}, who devised a set of features that include sentence position, common/uncommon keywords, proper nouns, numerical data, and sentence length. Most importantly, they devised a graph-based method to measure inter-sentence similarity and incorporate it as a feature. They also introduced the use of mathematical regression to estimate the weights of each feature. \cite{aristoteles2012text} applied a similar approach to text in the Indonesian language. 

The most recent approach was by \cite{Garc}; they applied a pre-processing step to find related words before using traditional summarization methods. They introduce several new parameters when generating summaries, such as maximum word counts. \cite{Garc} differed from previous work in that they used a heuristic (sentence position) to guide summarization, which was accomplished via word selection. They also encode target summarization in a manner dependent on a user's target summary length, requiring a custom implementation of uniform crossover.

Our work differs from that of \cite{fattah2008automatic} and \cite{Garc} as we want to explore a generalizable method relies more on semantics rather than paragraph composition. As mentioned previously, both of these work use an extensive amount of sentence-level heuristics to guide summarization. These heuristics make the trained model much less generalizable to unseen datasets, particularly those that may be in different formats, languages, or domains. 

\section{Motivation}
Current NLP work utilizes in deep learning, which creates computationally expensive models unsuitable for low-power devices such as mobile phones. We choose to explore GA-based models due to their customizability, easily allowing for feature selection and solution representation optimization. Although some text summarization work with GAs has been done (\cite{Garc} and \cite{fattah2008automatic}), these experiments were done several years prior to the significant advances in the NLP field in the later 2010's. Since the publication of these works, numerous pre-processing techniques that can be used to improve results of NLP tasks, such as Byte-Pair Encoding \cite{sennrich2016neural}, have been introduced. We examine the results of using a Genetic Algorithm for summarization as an alternative to deep learning, hypothesizing that a GA would be able to learn some set of generalizable word weights that can be used to perform summarization, the representation of which could be constructed to better adapt to low-power devices.
\section{Methodology}

\subsection{Summarization}
Our approach is done in the following steps. We first parse the entire training set, building a vocabulary of words to be used. When parsing each sample, we first tokenize the data using Moses Tokenizer \cite{koehn-etal-2007-moses}. Tokenization splits each line into tokens, which can either be a word or a symbol, creating the basis of our vocabulary. All words are converted to lower-case to reduce vocabulary size.  

We represent each individual in the GA's population as an array of weights ranging between 0 and 1 (inclusive). Each weight corresponds to a word/symbol in the vocabulary. To perform summarization of a sample in the dataset, we parse each line in the sample. The weight of each sentence is the average of the weights of each word in the sentence. We default unknown tokens (those that are not in the vocabulary) to a weight of zero. Only sentences whose weight passes some hyper-parameter threshold are included in the summarization.

\subsection{Evaluation}
We evaluate the summarization of any given sample using ROUGE-1 score \cite{lin-2004-rouge}. ROUGE is the standard evaluation metric for text summarization tasks, and has been showed to be correlated to the gold standard of human judgement \cite{lin2003automatic}. We use ROUGE-N specifically, which evaluates based off of n-gram alignment between the target and predicted summarization. There are three parts of a ROUGE score: F1, precision, and recall. Precision is defined as the number of true positive answers over the sum of the number of true positives and false positives. Recall is number of true positives over the sum of true positives and false negatives. F1 is the mean of precision and recall. In a more context specific manner, precision refers to the percent of n-grams in the target that are also in the prediction. Recall refers to the percent of n-grams in the prediction that are also in the target. 

\subsection{Dataset}
We use the CNN / Daily Mail dataset \cite{See:point} for both training and testing. The CNN portion itself contains over 90,000 article and summary pairs, which we found to be more than enough for the purposes of our experiments. We split the dataset into training and testing sets in different ways, which will be highlighted later on, to test how the GA is affected by the changes

\subsection{Genetic Algorithm}
There are several parameters of the GA that we experiment with and others that we leave constant throughout all of the experiments. This section will outline both categories in detail. 

\subsubsection{Population} We keep a constant population size of 100 throughout all of the experiments. This parameter, along with vocabulary size, was heavily influential on the GA's runtime and thus made experiments with larger population sizes quite costly. Since we were more interested in how the GA responded to changes in the problem representation itself, we did not feel the need to study this parameter extensively. 

As mentioned earlier, we represent each individual in the population as an array of weights, each of which correspond to a token in the vocabulary. Individuals in the population are initialized to random weights. 

\subsubsection{Fitness Function} We evaluate each individual with a ROUGE-1-n score averaged across all samples in the dataset, ranking for maximization. The score of each sample in the dataset is the average of the three ROUGE sub-metrics: F1, precision, and recall.

\subsubsection{Crossover} Our problem representation did not require any novel crossover operators, so we conduct our experiments using two-point crossover with a crossover rate of 0.8. We leave any experimentation with crossover operators to future work. 

\subsubsection{Mutation} We identified the mutation operator as a point of interest due to its ability to alter local search spaces. Mutation can be implemented in several different ways for this problem representation. We experiment with deletion mutation (set the weight of the gene to 0) with a  mutation rate of 0.01. Alternative mutation operators were considered, but further research was ultimately deemed out of the scope of the project.

\subsubsection{Selection} We keep tournament selection with tournament size of 5 as a constant parameter throughout all experiments. 

\subsubsection{Generations} We choose to limit the GA for 15 generations in each experiment, again due to time constraints. 

\section{Experiments and Results}

\subsection{Experimental Set up}
We implement our experiments in Python, due to the extensive NLP libraries available to the language, such as the Moses Tokenizer \cite{koehn-etal-2007-moses} and ROUGE \cite{lin-2004-rouge}. 
We create our genetic algorithms using the DEAP \cite{DEAP_JMLR2012} framework due to its large amount of documentation and support relative to other evolutionary computation frameworks. 

We conduct several experiments, altering the size of the training set and the portion of the training set used to generate the vocabulary. Due to limited time and computational power, we had to keep parameters value low. We run trials with training set sizes of 50 and 100. Vocabulary dataset size varied from 50, 1000, to the entire set (90,000). The sentence weight threshold remains constant in every experiment at a value of 0.6.

\subsection{Results}
Table \ref{table:rouge} compares the fitness ROUGE score (higher is better) of the best individuals found by a GA in each training set size and vocabulary set size pair. One particular point of interest is the somewhat lack of change in ROUGE score between vocabulary set sizes. We hypothesize that this may be because the GA has been able to learn how to determine sentence importance based off only the most common words used in the English language that are commonly shared across all articles in the dataset.

Table \ref{table:rougeTest} compares the results of the trained models on the testing set, which was comprised of 50 articles. Unsurprisingly, a larger training set and vocabulary size generally correlated with higher ROUGE scores. A notable exception was the score of the model trained on 50 samples with a vocabulary built from those 50 samples. That model outperformed one that was trained on 100 samples instead and another that had a vocabulary built from 1,000 samples. We hypothesize that this may be because of a better "match" between training data and vocabulary, indicating that the GA is able to better learn when given semantically appropriate context.

\begin{figure}
    \centering
    \includegraphics[width=9cm]{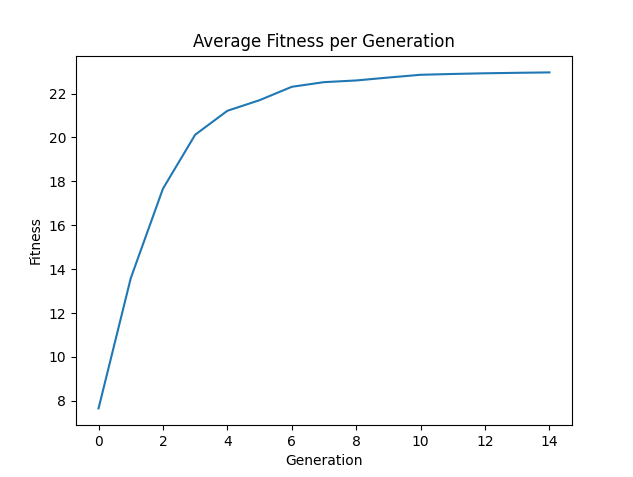}
    \caption{Average fitness of the population in each generation of a GA trained on 100 training samples and a vocabulary built from 90,000 files.}
    \label{fig:100_90k_average}
\end{figure}

\begin{figure}
    \centering
    \includegraphics[width=9cm]{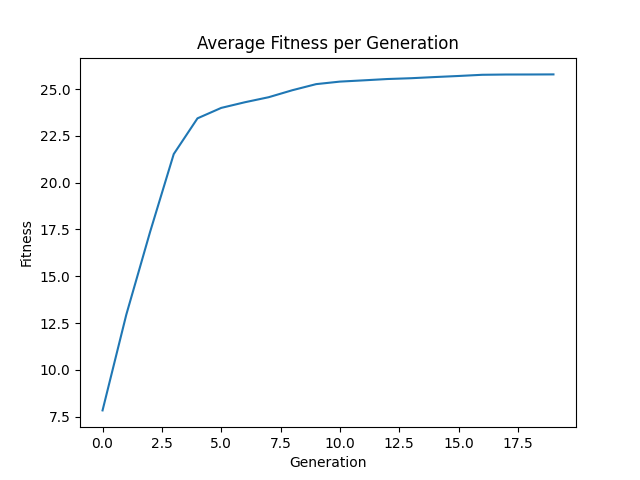}
    \caption{Average fitness of the population in each generation of a GA trained on 100 training samples and a vocabulary built from 1,000 files.}
    \label{fig:100_1000_average}
\end{figure}

Figure \ref{fig:100_90k_average} graphs the average fitness of the population in each generation for a GA that was trained on 100 articles and used a vocabulary built from the entire dataset. The chart shows that the GA quickly converges to a local optima well before it reaches the generation limit. There are approximately 35,000 tokens in the vocabulary set generated from 1000 files and about 325,000 tokens in the vocabulary tokens in the vocabulary set generated from 90,000 files. We suspected that this may be because of the extremely large solution space relative to the population size of only 100 individuals. Results (Figure \ref{fig:100_1000_average}) using a vocabulary of 35,000 tokens show the same phenomena, as do models trained using smaller training and vocabulary sizes (not depicted). 

Another noteworthy observation is that the GA is able to learn a small problem set when given a large problem representation. The fact that the GA can obtain better performance with an excessive vocabulary size compared to an adequate vocabulary size shows that it is able to filter out any "noise" in the solution space and focus specifically on the words of interest.

\begin{table}[htb]
  \centering
\begin{tabular}{|c|c|c|}
  \hline
  Training Size & Vocab Size &  ROUGE Score \\
  \hline
  100 & 90,000 & 28.6 \\ 
  \hline
  100 & 1,000 & 25.78 \\ 
  \hline
  100 & 50 & 23.98 \\ 
  \hline
  50 & 90,000 & 25.03 \\ 
  \hline
  50 & 1,000 & 26.39 \\ 
  \hline
  50 & 50 & 26.26 \\ 
  \hline
\end{tabular}
\caption{ROUGE scores of Best Performing Individuals on Training Set}
\label{table:rouge}
\end{table}

\begin{table}[htb]
  \centering
\begin{tabular}{|c|c|c|}
  \hline
  Training Size & Vocab Size &  ROUGE Score \\
  \hline
  100 & 90,000 & 23.59 \\ 
  \hline
  100 & 1,000 & 20.6 \\ 
  \hline
  100 & 50 & 18.88 \\ 
  \hline
  50 & 90,000 & 22.6 \\ 
  \hline
  50 & 1,000 & 18.47 \\ 
  \hline
  50 & 50 & 19.25 \\ 
  
  \hline
\end{tabular}
\caption{ROUGE scores of Best Performing Individuals on 50 sample Testing Set}
\label{table:rougeTest}
\end{table}

\section{Conclusion}
State of the art deep learning models require large amounts of computational power, even when only using a trained model with a relatively small amount of parameters. Previous works relied upon sentence heuristics, making them less generalizable across formats. We introduce the use of a genetic algorithm trained upon the semantics of sentences to create a simple text summarization model. Our experiments showed that the genetic algorithm is able to learn such features by being able to effectively navigate a large search space of a smaller problem representation.

\section{Future Work}
The implementation of the mutation operator serves as a point of interest for future exploration. Our experiments implement mutation as gene deletion by setting the weight of the gene (ie. the corresponding token in the vocabulary) to zero. Another possible method would simply be using a new random weight. Future experiments would benefit from an increased population size. Our experiments showed that the GA converged extremely quickly with a population size of 100, indicating that it was not sufficient to maintain diversity within the population. The increase in population size would likely require an increase in the number of generations for individuals to be evaluated as well. Finally, we suspect that it would be beneficial to encode the sentence weight threshold as part of the chromosome. This would make the GA far less reliant on hyper-parameters while allowing it to better generalize across different datasets.
\bibliographystyle{named}
\bibliography{ijcai21}
\end{document}